\documentclass[runningheads]{llncs}
\usepackage{graphicx}
\usepackage{longtable}
\usepackage{tabularx}
\usepackage{booktabs}
\usepackage{siunitx}
\begin{document}
\title{Teamwork Dimensions Classification Using BERT}
%
%
\author{Junyoung Lee\inst{1}\orcidID{0000-0001-6454-2990} \and
Elizabeth Koh\inst{2}\orcidID{0000-0002-2808-8687}}

\authorrunning{J Lee and E Koh}

\institute{Nanyang Technological University, Singapore\\
\email{junyoung002@e.ntu.edu.sg}\\
\and
National Institute of Education, Nanyang Technological University, Singapore\\
\email{elizabeth.koh@nie.edu.sg}}
\maketitle              
\begin{abstract}
Teamwork is a necessary competency for students that is often inadequately assessed. Towards providing a formative assessment of student teamwork, an automated natural language processing approach was developed to identify teamwork dimensions of students’ online team chat. Developments in the field of natural language processing and artificial intelligence have resulted in advanced deep transfer learning approaches namely the Bidirectional Encoder Representations from Transformers (BERT) model that allow for more in-depth understanding of the context of the text. While traditional machine learning algorithms were used in the previous work for the automatic classification of chat messages into the different teamwork dimensions, our findings have shown that classifiers based on the pre-trained language model BERT provides improved classification performance, as well as much potential for generalizability in the language use of varying team chat contexts and team member demographics. This model will contribute towards an enhanced learning analytics tool for teamwork assessment and feedback. 

\end{abstract}
\section{Introduction}
Teamwork is a necessary competency for students in the current education system in preparation for today’s global and complex environment, as well as the working world. Through collaboration with others, students can contribute their own knowledge and also learn from others. Analysis of the language used in discussions could shed light on aspects of behavior that the students might be unaware of during this collaborative process, and help them understand their roles and contributions in the setting. Moreover, teamwork is often inadequately assessed in schools as it can be complex and difficult when there are many teams. Analysis of the teamwork competencies of students can also complement teachers with an insight to the discussion process, and it can also provide formative assessment to students to help them improve their teamwork.

The framework used to identify teamwork competencies from students' dialogue was derived from \cite{koh2018}. This work focuses on the following four dimensions:
\begin{itemize}
    \item \textbf{Coordination (COD)}: organizing timely completion of task;
    \item \textbf{Mutual performance monitoring (MPM)}: keeping performance of other members in check;
    \item \textbf{Constructive conflict (CCF)}: resolving and diffusing adversarial situations between members via discussion and clarification; and
    \item \textbf{Team emotional support (TES)}: supporting members emotionally and psychologically.
\end{itemize}

An extension of the work in \cite{koh2018} was to identify teamwork dimensions from online chats of students, to provide a formative assessment of teamwork. Previous works on automatic analysis of teamwork dimensions on chat text have focused on rule-based classification \cite{shibani2017}, which requires labor-intensive rule writing, and machine learning-based classification \cite{shibani2017,suresh2018}, which raised concerns of generalizability to new data. \cite{suresh2018} had also proposed feature engineering as a way to improve classification performance.

Developments in the field of natural language processing and artificial intelligence resulted in novel approaches namely pre-trained language models (PLMs), such as the Bidirectional Encoder Representations from Transformers (BERT) model \cite{bert2018}, that allow for more in-depth understanding of the context of the text through attention models. Other downstream tasks in the field of artificial intelligence in education, such as automated short answer grading or assessing learner sentiments from online discussion forums, have also been successfully supplemented with PLMs \cite{moocbert2020,autograding2020}.

This work aims to build upon previous works in improving the classification performance of chat messages into teamwork dimensions, and seeks to explore the applicability of a BERT-based classifier over traditional machine learning models. Furthermore, generalizability of the proposed model would be evaluated on unseen data, as it would be ideal in developing a teamwork analytics system for learning groups of different age groups and demographics, especially with regard to the resource-consuming nature of manually annotating a large corpus.

\section{Experimental Setup}
\subsection{Corpus}
The corpus consists of chatlog gathered from 76 teams of a total of 272 14- year-old students in a Singapore secondary school, as part of a larger research study on teamwork. The students were given approximately 45 minutes for an ice-breaking activity, in which they were asked to describe their ideal teacher, followed by a dilemma task regarding environmental conservation. The members of the team were physically separated to prevent verbal communication, and the team discussions regarding the activity were carried out via an online chat system.

A total of 19762 raw chat messages were gathered from the participants, inclusive of spam lines. A portion of the chat messages from 7 teams was labeled independently according to the aforementioned teamwork dimensions by both of the two annotators, with Cohen's Kappa $\approx$ 0.65, indicating substantial inter-rater agreement. The rest of the dataset was divided and labeled individually by each of the annotators. Each message could be annotated for any number or none of the four teamwork dimensions. The number of non-disjoint positive labels for each category is as follows: COD — 2653; MPM — 1357; CCF — 2980; TES — 3506. A sample of the annotated dataset is shown in Table 1.


\begin{table}
\caption{Sample of the annotated dataset}\label{tab1}
\begin{tabularx}{\textwidth}{lXcccc}
\toprule
User &  Message & COD & MPM & CCF & TES\\
\midrule
Student A &  Bob are you okay with it & 1 & 1 & 0 & 0\\
\midrule
Student B &  ideal teacher as like um can work well with students and listen to students ideas & 0 & 0 & 1 & 0\\
\midrule
Student C & yes & 0 & 0 & 0 & 1\\
\midrule
Student C & So, caring, attentive to our needs and humourous? & 0 & 0 & 1 & 0\\
\midrule
Student A & humour yes & 0 & 0 & 0 & 1\\
\midrule
Student B & ok & 0 & 0 & 0 & 1\\
\bottomrule
\end{tabularx}
\vspace{-5mm}
\end{table}

\subsubsection{Text Pre-processing}
The chat messages were then pre-processed according to the steps outlined in Table 2, as proposed in \cite{shibani2017} to simplify the text. A challenge in the analysis of the chatlogs for teamwork competencies comes from prevalence of \textit{Singlish}, a colloquial English spoken in Singapore, as well as textese in chat message environment among youths \cite{weng2017}. Hence, pre-processing also served to improve compatibility with conventional English vectorizers. 

\begin{table}
\caption{Example of text pre-processing rules}\label{tab2}
\centering
\begin{tabularx}{\textwidth}{
    X
    >{\ttfamily}l
    @{\quad$\longrightarrow$\quad}
    >{\ttfamily}l
}
\toprule
 Procedure &  \textrm{Raw text} & \textrm{Preprocessed text}\\
\midrule
Emotion \& punctuation tagging 
    & ? & \{\{question\_mark\}\}\\
    & :) & \{\{pos\_emo\}\}\\
\midrule
Abbreviation expansion 
    & ikr & I know right \\
    & omg  & oh my goodness \\
\midrule
Local terms replacement 
    & macam & similar \\ 
    & chim & difficult\\
\midrule
Named Entity Recognition & Bob & \{\{NAME\}\} \\
\bottomrule
\end{tabularx}
\end{table}

Table 3 summarizes the feature engineering carried out on the corpus. As proposed by \cite{suresh2018} to improve the chat message classification performance, features based on context-sensitive rules were created using indicative terms, part-of-speech tagging, and regular expressions.

\begin{table}[t]
\caption{Feature engineering carried out on the corpus \cite{suresh2018}}\label{tab3}
\centering
\begin{tabularx}{\textwidth}{
    >{\ttfamily}lXX
}
\toprule
 \textrm{Feature} & Description & Example\\
\midrule
F\_TIME  &  Messages that mention task-related time constraint & ``faster lah" \newline ``we have like 15 mins left"\\
\midrule
F\_INSTRUCTION & Messages that instruct other team members for the task & ``see the url" \newline ``guys can we discuss now" \\
\midrule
F\_PROGRESS & Messages that indicate their progress on the activity & ``we r done" \newline ``we completed"\\
\midrule
F\_ELABORATION &  Messages that elaborate on discussed ideas & ``plants dont reduce smoke" \newline ``the teacher should be kind" \\
\midrule
F\_GREETING & Messages that are greetings &``sup" \newline ``good morning guys"\\
\midrule
F\_POSEMO & Messages that express positive emotions or humor & ``just kidding"\newline ``LOL"\\
\midrule
F\_AGREEMENT & Messages that indicate agreement &``yes ok" \newline ``yes thats possible"\\
\bottomrule
\end{tabularx}
\end{table}

For the purpose of this work in assessing the generalizability of a BERT-based classifier, the classification performance on both versions of the corpus pre- and post-feature engineering have been evaluated.

\subsection{Model Training}
The pre-processed, feature-engineered corpus described above has been split in the 6:2:2 ratio for training, validation, and testing with a preset random seed. The model was set up using the PyTorch Lightning library with the \texttt{bert-base-cased} pre-trained model and tokenizer from the HuggingFace Transformers library \cite{wolf-etal-2020-transformers}. Maximum sequence length for the transformer was set to 200. A linear layer was added as a final layer for multilabel classification. AdamW optimizer \cite{Loshchilov2017DecoupledWD} was used with warm-up steps of one-third epoch to a learning rate of $2\times 10^{-5}$ and linear decay. The criterion was \texttt{BCEWithLogitsLoss}, which combines a sigmoid layer with binary cross-entropy loss for multilabel classification. The maximum epoch was set at 100. The model was trained in the Google Colab environment with 16 GB of RAM on a single Tesla T4 GPU.

\section{Results}
\subsection{Classification Performance}
Due to the imbalance in number of positive labels for each class, a comparison of absolute accuracies between models would not be an appropriate gauge for model performance. Instead, macro-averaged precision, recall, F1 score, as well as Hamming distance (also known as Hamming loss) have been calculated for comparison with the best-performing machine learning model proposed by \cite{suresh2018}, a random forest (RF) classifier with term frequency-inverse document frequency (TF-IDF) vectorizer. Table 4 reports the results.

\begin{table}[!htbp]
\caption{Classification performance comparison}\label{tab4}
\centering
\begin{tabularx}{\textwidth}{
    l
    S[table-format=1.3]
    S[table-format=1.3]
    S[table-format=1.3]
    S[table-format=1.3]
}
\midrule
 &\multicolumn{2}{c}{Without feature engineering}& \multicolumn{2}{c}{With feature engineering}\\
\midrule
& {RF} & {BERT} & {RF} & {BERT} \\
\midrule
Precision & \bfseries{0.824} & 0.810 & \bfseries{0.833} & 0.801\\
Recall & 0.406 & \bfseries{0.702} & 0.594 & \bfseries{0.747}\\
F1 Score & 0.527 & \bfseries{0.734} & 0.684 & \bfseries{0.757} \\
\midrule
Hamming distance & 0.124 & \bfseries{0.076} & 0.090 & \bfseries{0.070} \\
\midrule
\end{tabularx}
\end{table}

On test sets with or without additional features, the BERT-based classifier shows comparable precision score to the RF classifier, while outperforming in the other three metrics. The BERT-based classifier also shows nearly consistent performance on both versions of the test data, while the RF classifier displays greater degradation when tested on data without engineered features.

\subsection{Inference on Unseen Data}
To assess the generalizability of the trained model, it is beneficial to evaluate on a set of unseen data which differs in participant demographic, language use, and task context. In a separate setting, 2 groups of 15 students aged 19 to 42 have been invited to utilize the online chat system to brainstorm for a community service project as an in-class activity. An anonymized chatlog of 129 messages in total was collected, and manually annotated with the four teamwork dimensions. It was pre-processed using the steps in Table 2 with no additional features. The trained BERT-based classifier achieved a Cohen's Kappa score of \textbf{0.640} compared to 0.467 achieved by the RF classifier, showing a higher inter-rater reliability.

\section{Conclusion}
In this work, multilabel classification performance of a BERT-based classifier was evaluated on both seen and unseen corpus of chat messages generated by students discussing a team task, to analyze their teamwork competencies. The results show much potential for scalability of such models without need for extensive feature engineering, and their generalizability with different demographics, in building a reliable teamwork analytics system from online chats. This improved classifier model ultimately contributes towards an enhanced formative learning analytics tool for teamwork assessment, allowing students and teachers to become more aware of the teamwork competencies in the discussion process and receive more holistic feedback.

\section*{Acknowledgements}
This paper refers to data and analysis from OER62/12EK and OER09/15EK, funded by the Education Research Funding Programme, National Institute of Education (NIE), Nanyang Technological University, Singapore and the Incentivising ICT use Innovation Grant (I3G 9/19EK) from NIE. IRB: IRB-2020-08-006.

%
%
\bibliographystyle{splncs04}
\bibliography{refs}

\end{document}